\documentclass[journal]{IEEEtran}

\usepackage{bm}
\usepackage{multirow}

\usepackage{comment}
\usepackage{color}
\usepackage{amssymb}
\usepackage{hyperref}
\usepackage[hyphenbreaks]{breakurl}
\usepackage{cite}

\ifCLASSINFOpdf
\else
\fi
\usepackage{amsmath}
\usepackage{algorithmic}
\usepackage[ruled,linesnumbered]{algorithm2e}

\usepackage{array}

\usepackage[caption=false,font=footnotesize]{subfig}
\usepackage{graphicx}
\usepackage{fixltx2e}

\usepackage{stfloats}
\usepackage{comment}

\usepackage{url}

\begin{document}
%
\title{Prototype-Guided Memory Replay for \\ Continual Learning}
%
%
%

\author{Stella~Ho,
        Ming~Liu,
        Lan~Du,
        Longxiang~Gao 
        and Yong~Xiang 
\IEEEcompsocitemizethanks{\IEEEcompsocthanksitem S. Ho, M. Liu, and Y. Xiang are with the School of Information Technology, Deakin University, Geelong, VIC 3220, Australia.\protect\\
E-mail: \{hoste, m.liu, yong.xiang\}@deakin.edu.au
\IEEEcompsocthanksitem Corresponding author: L. Gao is with the School of Information Technology, Deakin University, Geelong, VIC 3220, Australia.\protect\\
E-mail: longxiang.gao@deakin.edu.au
\IEEEcompsocthanksitem L. Du is with the Faculty of Information Technology, Monash University, Clayton, VIC 3800, Australia.\\
Email: lan.du@monash.edu}
\thanks{This work has been submitted to the IEEE for possible publication. Copyright may be transferred without notice, after which this version may no longer be accessible. This is the pre-print version which has not been fully undergone peer review.}}

\maketitle

\begin{abstract}
Continual learning (CL) refers to a machine learning paradigm that learns continuously without forgetting previously acquired knowledge. Thereby, major difficulty in CL is catastrophic forgetting of preceding tasks, caused by shifts in data distributions. Existing CL models often save a large number of old examples and stochastically revisit previously seen data to retain old knowledge. However, the occupied memory size keeps enlarging along with accumulating seen data. Hereby, we propose a memory-efficient CL method by storing a few samples to achieve good performance. We devise a dynamic prototype-guided memory replay module and incorporate it into an online meta-learning model. We conduct extensive experiments on text classification and investigate the effect of training set orders on CL model performance. The experimental results testify the superiority of our method in terms of forgetting mitigation and efficiency.
\end{abstract}


%
\IEEEpeerreviewmaketitle

\section{Introduction}
\IEEEPARstart{C}{urrent} dominant machine learning paradigm uses samples from an identically and independently distribution (i.i.d) to train a leaner. It often needs a large amount of training samples to ensure model performance. In practice, many learning tasks have a large volume of unlabelled data. High quality training samples may become available through time, especially for tasks from new emerged domains. Such that, conventional learning models have to learn from scratch each time when new training samples becomes available \cite{DBLP:journals/corr/abs-1909-08383}. In particular, these models have to accumulate training samples to improve their learning abilities. However, previously learned examples may become inaccessible through time due to privacy reasons. Continual Learning (CL) is a solution to cope with dynamic data distribution and address these problems \cite{dAutume2019EpisodicMI,LopezPaz2017GradientEM}.

Intuitively, CL simulates human learning process, in which human learn novel information sequentially, and often revisit old knowledge to enhance their learning ability for new information \cite{DBLP:journals/corr/abs-1909-08383}. The major research challenge in CL is to address inadequate learning performance caused by catastrophic forgetting. Catastrophic forgetting \cite{McCloskey1989CatastrophicII} is a \textit{stability-plasticity} dilemma. \textit{Stability} refers to preservation of previous knowledge. While, \textit{plasticity} refers to integrating novel knowledge. We expect a CL model to continuously learn novel information while preserving the previously acquired knowledge. A CL model does not need to learn from scratch to adapt to new tasks.

Existing CL models can be categorized into three mainstream methods i.e., regularization-based methods \cite{Kirkpatrick2017OvercomingCF,Li2018LearningWF,2017Continual}, architecture-based methods \cite{yoon2018lifelong,DBLP:conf/iclr/Adel0T20, 9724647,9181489} and replay-based methods \cite{Wang_2019,dAutume2019EpisodicMI,wang-etal-2020-efficient,AGEM,sun2020lamal}. Regularization-based method typically limits the updates of non-trivial parameters or weights by adding constraints. Due to the complexity of deep learning models, a small change in parameter space could have unexpected effects on the model's performance \cite{Wang_2019}. Architecture-based method dynamically expands model architecture to learn new tasks. It preserves previously finetuned parameters and introduces task-specific parameters for new tasks. However, the size of the parameters grow as the number of seen tasks increases. Most existing CL models are replay-based methods. Replay-based method revisits past data or generates pseudo past data to consolidate knowledge. However, most replay-based models neglect the memory constraint in the CL setup. The size of the saved samples grows aggressively as more tasks are seen. Hence, it is imperative to efficiently save and use samples for memory replay.

This paper introduces a prototype-guided memory replay module, which dynamically updating learned knowledge. We employ a prototypical network \cite{Snell2017PrototypicalNF} as a feature extractor to generate a prototype for each learned class where each prototype serves as the delegate of a learned class. Prototypical knowledge is used for guiding samples selection for sparse memory replay. We also apply meta learning to enable few-shot learning on novel tasks and knowledge transfer among tasks. In the experiments, we propose a low-resource CL setup, where we strictly limit the number of stored samples to a fixed value. Model performance is evaluated on text classification in terms of accuracy in different training set orders. We conduct a further investigation on the effect of training set orders on CL model performances. Additionally, ablation study is performed on replay sample selection strategies. Extensive experimental outcomes on Yelp, AGNews and Amazon datasets \cite{c9d4fbeac7324056bed5d1cb262a7268} testify the superiority of the proposed model as a CL leaner.

Our main contributions are threefold and summarized as follows.
\begin{itemize}
    \item We propose a continual learning method, namely PMR, that incorporates an online meta-learning model and a prototypical network. The proposed method is able to consolidate knowledge by only revisiting a few examples. 
    
    \item We devise a prototype-guided memory replay network, where prototypes are constantly corrected in the training and and guide sample selection to achieve sample efficiency. 
    
    \item In experiments, we manifest that the performance of CL models severely depends on the order of training sets if data distribution differs a lot.
    
\end{itemize}

The remainder of this paper is organised as follows. We present the related work in Section \ref{sec:related work}. We then introduce the problem formulation of CL in Section \ref{sec: problem formulation}. Section \ref{sec:cl-pmr} depicts the proposed approach with an illustration of the model and detailed algorithms. In Section \ref{sec:experiments}, extensive experimental outcomes are thoroughly discussed and analyzed. Section \ref{sec:conclusion} concludes the paper.

\section{Related Work} \label{sec:related work}
\subsection{Replay-based Methods}
Replay-based methods store certain amount of the previous training data and episodic replay them with the novel training data to ensure knowledge consolidation. Replay-based approaches focus on employing a proper sample selection scheme for memory replay, so that it could better address stability-plasticity dilemma. Gradient Episodic Memory (GEM) \cite{LopezPaz2017GradientEM} alleviates catastrophic forgetting via restricting gradient projection. It performs memory replay by gradient computations from memory data and mitigates the training loss on previously learned tasks by gradient constraint. Due to the complexity of distribution approximation problem, GEM utilises a simply selection scheme for replay, i.e., it randomly stores samples from each task. But, GEM fails to address the issue of time complexity \cite{Wang_2019}. A(verge)-GEM \cite{AGEM} is a simple and efficient variant of GEM, where only projects on one direction outlined by randomly chosen examples from a preceding task data buffer. The projected gradient is imposed to diminish the average training loss on previously learned tasks. 

Many replay-based approaches utilise episodic memory replay with a 1 \% sparse replay rate, which was first proposed in MbPA++ by \cite{dAutume2019EpisodicMI}. Nevertheless, in its inference process, MbPA++ \cite{dAutume2019EpisodicMI} retrieved k nearest neighbours using Euclidean distance function for local adaptation, which is expensive. To address this problem, \cite{DBLP:journals/corr/abs-2009-04891} introduced a replay frequency to harness the replay sparseness in terms of time and size. However, they neglected memory constraints in CL by writing all seen samples into memory in the implementation. The size of  occupied memory was explosively expanding with the increase of training samples seen by the model.

Other works \cite{Shin2017ContinualLW,sun2020lamal,kemker2018fearnet} were considered as generative replay-based approaches, which use generative models rather than memory modules. LAMOL \cite{sun2020lamal} applied generative models to generate pseudo-samples from previous tasks for experience replay.

\subsubsection{Sample Selection Schemes}
Most replay-based approaches randomly select data and save into memory module for experience replay later. For task-specific learning process, some approaches \cite{dAutume2019EpisodicMI,obamuyide-vlachos-2019-meta} applied $k$-center method to choose k data points according to their distances to centroid or other geometric properties. Both random selection and $k$-center methods utilise heuristics to update the memory. EA-EMR \cite{Wang_2019} and IDBR \cite{Huang2021ContinualLF} selected informative samples by referencing the centroid of the cluster via K-Means. iCaRL \cite{Rebuffi2017iCaRLIC} chose samples, that are nearest to the mean of the distribution. \cite{Lee2019OvercomingCF} employed short-term memory (STM) as a memory buffer to collect a sufficient newly seen data so that it could prevent over-fitting and bias toward the current learning task. \cite{wang-etal-2020-efficient} applied two popular paradigms in active learning into memory selection rules, i.e., diversity-based methods \cite{10.1007/978-3-540-74958-5_14} and uncertainty-based methods \cite{ramalho2018adaptive,toneva2018an}. However, random selection outperforms both of these two methods in terms of accuracy and efficiency.

\subsection{Regularization-based Methods}
Regularization-based approaches diminish the update of the important parameters or the neural weights to consolidate knowledge. Learning without forgetting (LwF) approach \cite{Li2018LearningWF} exerted knowledge distillation to retain the
representations of previous data from changing too much while learning novel tasks. However, LwF was weak of responding domain shift between tasks. A recent developed LwF variant \cite{8953661} with less-forget constraint performed regularization on the cosine similarity between the L2 Norm logits of the previously learned and current model. The elastic weight consolidation (EWC) was introduced by \cite{Kirkpatrick2017OvercomingCF}. It applied the Fisher Information Matrix to compute a diagonal approximation, and imposed a quadratic penalty on the changes of important neural network parameters. But, EWC failed to learn new classes incrementally. Similarly, Memory Aware Synapses (MAS) \cite{Aljundi2018MemoryAS} also used a diagonal approximation by accumulating the magnitude of the gradients as the sensitivity of the learned function. The sensitivity served as an importance measure for each parameter of the network. \cite{2017Continual} proposed a CL approach, which alleviate forgetting by penalizing the updates to the most relevant synapses. In particular, it computed synaptic relevance and operates in the parameter space. Additionally, variational continual learning \cite{v.2018variational} under the Bayesian-based framework was also widely used in recent regularization-based approaches \cite{Adel2020Continual,Ahn2019UncertaintybasedCL,Zeno2020TaskAC,DBLP:journals/corr/abs-1902-06494}. Recently, PCL \cite{DBLP:conf/aaai/HuQWML21} was proposed as a per-class continual learning model, which regularized parameters and solved feature bias by introducing holistic regularization. While, IDBR \cite{Huang2021ContinualLF} leveraged L2 Norm to perform regularization in task-specific space to some extent, and constrain the updates in the task generic space.

\subsection{Continual Learning in NLP}
A majority of the language models used deep learning networks. Whereas, regularization-based approaches operate on models' parameter spaces or gradient spaces. Due to the complexity of deep neural networks, working on parameter space was no longer a proper approach to represent model behaviour. Therefore, in order to address NLP tasks, most CL approaches favoured replay-based methods or/and carefully impose regularization in the embedding space \cite{Wang_2019,Huang2021ContinualLF}. For language representation learning, EA-EMR was introduced as an alignment model to address the lifelong relation detection. For a minimal distortion of the embedding space, a simple linear transformation was applied on the top of the original embedding, bridging the newly learned embedding and the previous used embedding space. IDBR \cite{Huang2021ContinualLF}, as a text classifier, utilised information disentanglement to differentiate generic representations and task specific representations on hidden spaces, where only task specific representation perform a large degree of updates in training. S2S-CL \cite{Li2020Compositional} was devised for language instruction learning. It leveraged compositionality \cite{li-etal-2019-compositional} by dividing input into semantic representation and syntactic representation, where  syntactic representation remains in training.

\subsection{Meta Learning}
Recently, meta learning was introduced in CL models, mainly in replay-based methods. Some CL approaches \cite{riemer2018learning,DBLP:journals/corr/abs-2009-04891,obamuyide-vlachos-2019-meta} showed that meta learning enables efficient positive forward knowledge transfer and diminishes interference between tasks. Meta-Experience Replay (MER) \cite{riemer2018learning} utilised optimization based meta-learning, Reptile \cite{Nichol2018OnFM} to regularize the objective of experience replay, but with a high replay rate. MLLRE \cite{obamuyide-vlachos-2019-meta} adopted the Reptile algorithm to meta updates for knowledge transfer to address lifelong relation extraction. Meta-MbPA \cite{wang-etal-2020-efficient} used meta-learning to find a good initialization point for sparse memory replay. While, OML-ER and ANML-ER \cite{DBLP:journals/corr/abs-2009-04891} utilised FOMAML \cite{Finn2017ModelAgnosticMF} to find a good initialization point for few-shot learning, augmented with sparse experience replay. OML-ER was the latest state-of-the-art CL model in text classification.   

\section{Problem Formulation} \label{sec: problem formulation}
We consider a basic CL setup, where models make a single pass over a stream of training examples. We assume that the training data stream consists of $K$ tasks, $\{\mathcal{T}^{(1)},\mathcal{T}^{(2)},...,\mathcal{T}^{(K)}\}$, in an ordered sequence. Each task $\mathcal{T}^{(k)}$ is a supervised learning task with a ground truth label set $D^{(k)}_{\mathrm{train}}=\{(x^{(k)}_{j}, y^{(k)}_{j})\}_{j=1}^{|D^{(k)}_{\mathrm{train}}|}$. The objective in a task $\mathcal{T}^{(k)}$ is to train a continual learner $f$ that converges to the probability distribution of $\mathcal{T}^{(k)}$. Meanwhile, we expect $f$ to perform well on previous tasks $\{\mathcal{T}^{(1)},\mathcal{T}^{(2)},...,\mathcal{T}^{(k-1)}\}$ without using the past training sets. The overall objective is to minimize the average expected risk of $K$ tasks seen so far as:
\begin{equation}
    \frac{1}{K} \sum_{k=1}^{K} \mathbb{E}_{x,y \sim p(\mathcal{T}^{(k)})}[\mathcal{L}(f_{\theta}(x),y)]
\end{equation} where $K$ denotes the total number of seen tasks and $\theta$ is the parameters of $f$. A replay-based method allows the model $f$ to preserve a certain amount of training samples from previous tasks. In a CL setup, we limit the memory size to a constant size $\mathrm{B}$, particularly the memory stores preceding seen samples with the amount less than or equals to $\mathrm{B}$.

CL has two research scenarios in text classification, i.e., class-incremental learning (CIL) and task incremental learning (TIL). In CIL, task identity information is not provided in test, and it has a single classifier for all classes. While, in TIL, task identity information is provided in both training and inference, and a CL model assigns one classifier for each task. Recent works mainly operate in a task-free setting \cite{8953745} where neither training nor test examples contain dataset identity. In order to fit a task-free setting, we address text classification tasks in a CIL scenario.

\subsection{Class-incremental Learning} 
CIL refers to class-incremental learning without task identity information provided in inference. Given an encoder $h_{\phi_{\mathrm{e}}}: \mathcal{X} \rightarrow \mathbb{R}^d$ and a classifier $g_{\phi_{\mathrm{pred}}}: \mathbb{R}^d \rightarrow \mathbb{R}^N$, where $d$ is the dimension of the hidden representations and N is the number of classes, we consider a model $f_{\theta} = g_{\phi_{\mathrm{pred}}}(h_{\phi_{\mathrm{e}}}(\boldsymbol{x}))$. The cross entropy loss for a task $\mathcal{T}^{(k)}$ is,
\begin{equation}\label{eqn:CE_task}
    \mathcal{L}_{CE}(\boldsymbol{x}, \boldsymbol{y};\theta^{(k)}) = - \sum^{|D^{(k)}_{\mathrm{train}}|}_{j=1} \sum^{N^{(k)}}_{i=1} y_{ji} \log (\sigma(f_{\theta^{(k)}}(x_j))_{i})
\end{equation} where $N^{(k)}$ is the number of classes in $\mathcal{T}^{(k)}$, $\sigma$ is the activation function(e.g., sigmoid or softmax) and  $(\boldsymbol{x}, \boldsymbol{y}) \in D^{(k)}_{\mathrm{train}}$. Alternatively, the cross entropy loss while learning a task $\mathcal{T}^{(k)}$ can be referred as the loss for all seen classes. For a replay-based methods, the cross entropy loss generally can be formulated as:

\begin{equation}\label{eqn:CE_ALL}
    \mathcal{L}_{CE}^{*}(\boldsymbol{x}, \boldsymbol{y};\theta^{(k)}) = - \sum^{|D^{(k)}_{\mathrm{train}}|+|\mathcal{M}|}_{j=1} \sum^{N}_{i=1} y_{ji} \log (\sigma(f_{\theta^{(k)}}(x_j))_{i})
\end{equation} where $\mathcal{M}$ denotes preceding seen data set that saved in Memory and  $(\boldsymbol{x}, \boldsymbol{y}) \in D^{(k)}_{\mathrm{train}} \cup \mathcal{M}$. In this paper, we mainly use the cross entropy loss shown in Eqn.\ref{eqn:CE_task}. More details are shown in Section \ref{subsec:meta-continual}.

\section{Continual Learning with Prototype-guided Memory Replay} \label{sec:cl-pmr}
Meta learning can realise efficient learning on new tasks via knowledge transfer in CL \cite{DBLP:journals/corr/abs-2009-04891}. Thus, we leverage meta learning with our prototype-guided memory replay network to formulate a CL model, called Prototype-guided Memory Replay model, or PMR.


\begin{figure}[!t]
\centering
\includegraphics[width=3.5in]{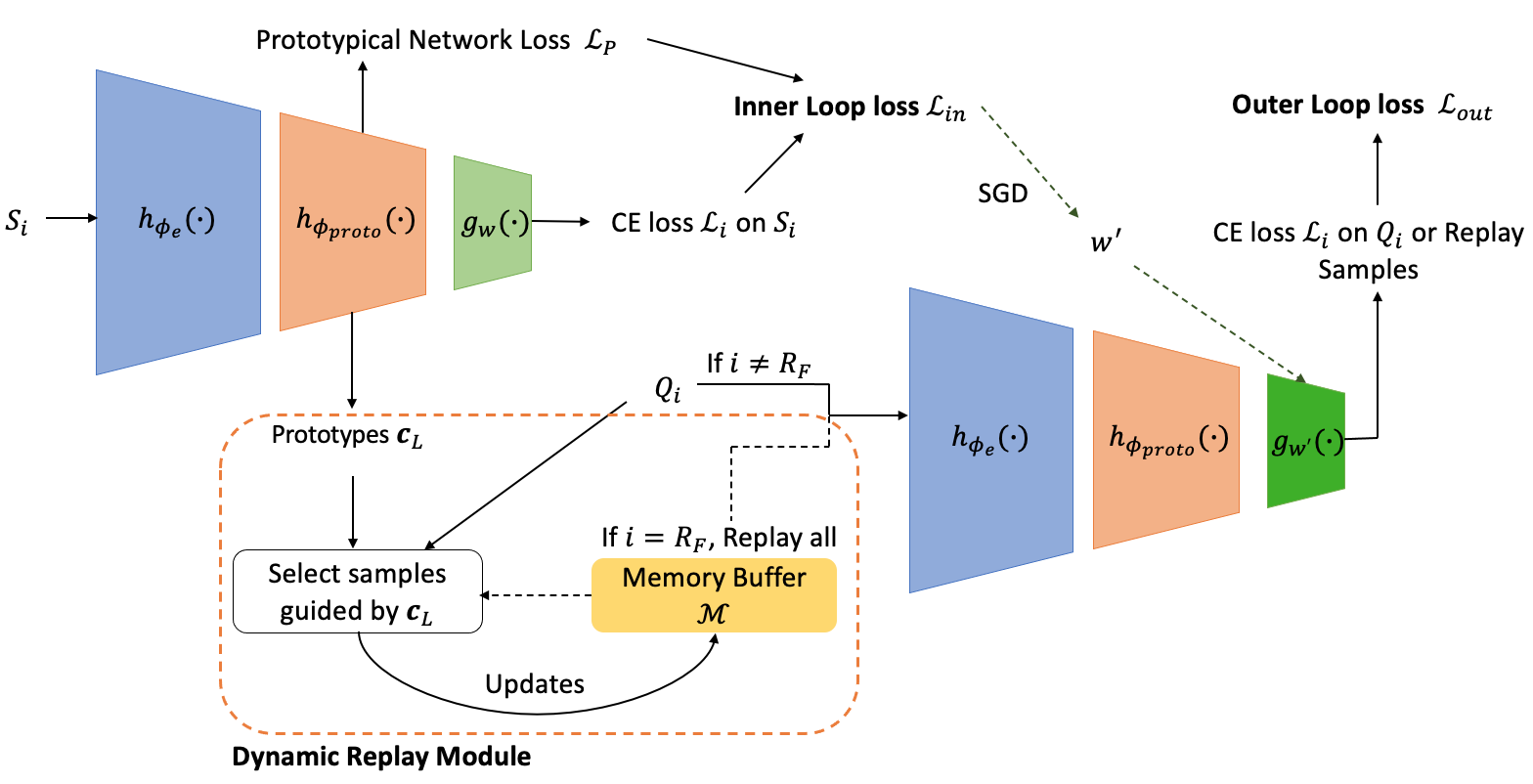}
\caption{An illustration of \textbf{PMR}. During training, samples in memory $\mathcal{M}$ are dynamic updating based on the accumulated knowledge representing by $\boldsymbol{c}_{\boldsymbol{K}}$. Inner loop loss $\mathcal{L}_{in}$ is calculated to update parameter $\boldsymbol{w}$. All parameters are meta learned through outer loop optimization on query set or replay samples.}
\label{fig:pmr-model}
\end{figure}

\subsection{Meta Continual Learning} \label{subsec:meta-continual}
Meta learning is a learning paradigm that enables quick adaptation from a novel domain. Recently, some CL models \cite{DBLP:journals/corr/abs-2009-04891, wang-etal-2020-efficient} utilise meta learning for domain adaptation and efficient knowledge transfer. Specifically, we expect to learn a good initialisation point so that the model could perform well on a novel task with merely a few gradient updates. The meta training process refers to a two-level optimization process. In an inner-loop optimization, the model applies task-specific finetuning on a support set. While, in the outer-loop optimization, we use a query set to perform meta-updates. Given parameters $\boldsymbol{\theta}$, model $f_{\boldsymbol{\theta}}$ and a task $\mathcal{T}^{(k)}$, the inner-loop optimisation is a $m-$step gradient-based update on the support set $S$, which results in parameters $\boldsymbol{\theta}'$. The outer loop optimization is for generalisation. We expect $f_{\boldsymbol{\theta}'}$ to well generalise over tasks from overall distribution $p(\mathcal{T})$. However, updating parameters $\boldsymbol{\theta}$ involves second-order gradients computation, which is computationally expensive. FOMAML \cite{Finn2017ModelAgnosticMF} provides a solution of this computational complexity issue and enables efficient knowledge transfer. Hereby, we employ FOMAML in our model's meta learning process.

\begin{table}[!t]
\renewcommand\arraystretch{1.3}
\centering
\caption{\label{notation}Notations}
\begin{tabular}{|c||c|}
\hline
\textbf{Notations} & \textbf{Explanation} \\
\hline
$k$ & Task index\\
\hline
$l$ & Class index\\
\hline
$i$ & Episode index\\
\hline
$D_{\mathrm{train}}$ & Training set \\
\hline
$N^{(k)}$ & Number of classes in task $\mathcal{T}^{(k)}$ \\
\hline
$f_{\boldsymbol{\theta}}$  & Model with parameters $\boldsymbol{\theta}$ \\
\hline
$h_{\boldsymbol{\phi}}$  &  Representation  learning  function with parameters $\boldsymbol{\phi}$\\
\hline 
$\boldsymbol{\theta}_{e}$ & Encoder's parameters \\
\hline 
$\boldsymbol{\theta}_{proto}$ & Prototypical network's parameters \\
\hline
$g_{\boldsymbol{w}} $ &  Prediction learning function with parameters $\boldsymbol{w}$ \\
\hline
$\mathcal{L}_{i} $ &  Cross entropy loss in episode $i$\\
\hline
$S_{i}$ &  Meta learning support set in episode $i$ \\
\hline
$Q_{i}$ &  meta learning query set in episode $i$ \\
\hline
$\mathcal{L}_{P} $ &   Loss function of prototypical network  \\ 
\hline
$S_{l}$ & Support set of class $l$ in prototypical network \\
\hline
$Q_{l}$ & Query set of class $l$ in prototypical network \\
\hline
$N_{S}$ & Number of support examples per class \\
\hline
$N_{Q}$ & Number of query examples per class \\
\hline
$\boldsymbol{c}_{l}$ & Prototype of class $l$ \\
\hline
$m$ &  Optimization step  \\
\hline
$R_F$ & Replay frequency \\
\hline
$J(\cdot) $ &  Objective function \\
\hline
$d(\cdot) $ &  Distance function \\
\hline
$\alpha$ &   Inner loop learning rate \\
\hline
$\beta$  &   Outer loop learning rate \\
\hline
\end{tabular}
\end{table}

\subsubsection{Online Meta Learning} The proposed model $f_{\boldsymbol{\theta}}$ consists of two learning networks. Given the representation learning function $h_{\boldsymbol{\phi}}$ with parameters $\boldsymbol{\phi}$ and the prediction learning function $g_{\boldsymbol{w}}$ with parameters $\boldsymbol{w}$, we express the function of the model with input $x$ as $f_{\boldsymbol{\theta}}(x) = g_{\boldsymbol{w}}(h_{\boldsymbol{\phi}}(x))$.

The representation learning network (RLN) consists an encoder and a prototypical learning network. In order to reduce model complexity and computational budgets, we employ ALBERT \cite{DBLP:journals/corr/abs-1909-11942} as our example encoder. $\boldsymbol{\phi}_e$ denotes the learnable parameters of our text encoder. We introduce a single-hidden-layer feed-forward neural network (NN) as our prototypes learning network. We denote the learnable parameters of the prototypical network as $\boldsymbol{\phi}_{proto}$. We use $\mathrm{ReLU}(\cdot)$ as the activation function of NN with dropout rate 0.2. More details regarding our prototypical network are elaborated in Sec.~\ref{subsec: proto-memory}. We use a single linear layer as the prediction learning network (PLN), in which output dimension equals to the number of classes.

Only $g_{\boldsymbol{w}}$ is fine-tuned in the inner loop optimisation. Given episode $i$, we perform SGD on the support set mini-batch $S_i$ to fine tune PLN and update parameters $\boldsymbol{w}$ as:

\begin{equation}
    \boldsymbol{w}' = \mathrm{SGD}( \mathcal{L}_{in}, \boldsymbol{\phi}, \boldsymbol{w}, S_i, \alpha)
\end{equation} And, the overall loss function of inner loop optimization is shown as: 

\begin{equation}
    \mathcal{L}_{in} = \mathcal{L}_{P}(\boldsymbol{\phi}_{proto}, Q_{L}) +  \mathcal{L}_i (\boldsymbol{\phi}, \boldsymbol{w}, S_i)
\end{equation} where $\mathcal{L}_{i}$ is the cross entropy loss on support set in episode $i$ and $\mathcal{L}_p$ is the loss of prototypical network on its query set $Q_{L}$. It is noteworthy that the query set $Q_{L}$ for prototypical network is different from that of meta learning. $Q_{L}$ is only used for prototypes computation while $Q_{i}$ is merely used for generalization in meta learning. In the outer loop optimisation, the parameters $\boldsymbol{\phi}$ and $\boldsymbol{w}$ are meta-learned using query set $Q_{i}$, that is:

\begin{equation}
    J(\boldsymbol{\theta}) = \mathcal{L}_i (\boldsymbol{\phi}, \boldsymbol{w}', Q_i) 
\end{equation} And the optimizer that we used is Adam optimizer \cite{Kingma2015AdamAM} with a outer loop learning rate $\beta$ as:

\begin{equation}
    \boldsymbol{\phi} \leftarrow \mathrm{Adam(J(\boldsymbol{\theta}), \beta)}
\end{equation}

The training and inference process of our model are described in Algorithm \ref{algo:training} and Algorithm \ref{algo:testing}, respectively. In Algorithm \ref{algo:training}, $\mathrm{RANDOMSAMPLE}(S, N)$ denotes a set of $N$ elements chosen uniformly at random from set $S$,  without replacement. $d(\cdot)$ denotes the distance function, where we leverage Euclidean distance to compute $\mathcal{L}_{P}$ in this paper.

\begin{algorithm}[t]
	\caption{Meta Training for PMR$_{argmin}$.}
	\label{algo:training}
	\KwIn{Initial model parameters $\boldsymbol{\theta} = \boldsymbol{\phi}_e \cup \boldsymbol{\phi}_{proto} \cup \boldsymbol{w} $, support set for episode $i$ $S_i$, support set buffer size $m$, replay frequency $R_F$, memory $ \mathcal{M}$, inner-loop learning rate $\alpha$, outer-loop learning rate $\beta$.}
	\KwOut{Trained model parameters $\boldsymbol{\theta}$, updated memory $\mathcal{M}$}
	\For{$i = 1, 2$, ...}{
	    $S_i \leftarrow m$ batches from the stream $D_{\mathrm{train}}$ \\
	    \For { \rm{class} $l$ \rm{in}  $S_i $} {
	        $S_l \leftarrow \mathrm{RANDOMSAMPLE}(S_{i,l}, N_S)$ \\
	        $Q_l \leftarrow \mathrm{RANDOMSAMPLE}(S_{i,l} \backslash S_l, N_Q )$ \\
	 	    $\mathbf{c}_l \leftarrow \frac{1}{|S_l|} \sum \limits_{\substack{ (x_i, y_i) \in S_l}} h_{\boldsymbol{\phi}_{proto}}(x_i)$ \\
		\For {$(x, y)$  \rm{in} $ Q_l $} { 
		   $\mathcal{L}_{P}({\boldsymbol{\phi}_{proto}}, Q_l) \leftarrow \mathcal{L}_{P}({\boldsymbol{\phi}_{proto}}, Q_l) + \frac{1}{N_Q}[d(h_{\boldsymbol{\phi}_{proto}}(\mathbf{x}), \mathbf{c}_l) + \log \sum \limits_{\substack{l'}} \mathrm{exp}(-d(h_{\boldsymbol{\phi}_{proto}}(\mathbf{x}), \mathbf{c}_l))]$}
	    }   
	    \eIf{$i = R_F$} { $Q_i \leftarrow$ sample$(\mathcal{M}, all) $  \tcp*{Sample all data points for each class from $\mathcal{M}$} 
	     }{  $Q_i \leftarrow $next batch from the stream \\
	    	\For{\rm{class} $l$ \rm{in} $Q_i $}{
	    	    writeSamples$(\mathcal{M}, Q_i, \boldsymbol{c}_l, n)$  \tcp*{Write $n$ data points for class $l$ into $\mathcal{M}$}
	    	}
		} 
	    $\mathcal{L}_{in} = \mathcal{L}_{P}({\boldsymbol{\phi}_{proto}}, Q_{L}) +  \mathcal{L}_{i} (\boldsymbol{\theta}, S_i) $  \\
	    $ \boldsymbol{w}' = \mathrm{SGD}( \mathcal{L}_{in}, \boldsymbol{\phi}_e, \boldsymbol{\phi}_{proto}, \boldsymbol{w}, S_i,  \alpha)$\\   
	    $J(\boldsymbol{\theta}) = \mathcal{L}_i (\boldsymbol{\phi}_e, \boldsymbol{\phi}_{proto}, \boldsymbol{w}' , Q_i)$ \\
	    $\boldsymbol{\theta} \leftarrow \mathrm{Adam(J(\boldsymbol{\theta}), \beta)} $ 
	    
	    \If{\rm{all the training data is seen}}{
	        \textbf{Stop Iteration}
	    }
	
	 }   
\end{algorithm}

\begin{algorithm}[t]
	\caption{Meta Inference for PMR$_{argmin}$.}
	\label{algo:testing}
	\KwIn{Initial model parameters  $\boldsymbol{\theta} = \boldsymbol{\phi}_e \cup \boldsymbol{\phi}_{proto} \cup \boldsymbol{w} $, support set buffer size $m$, memory $\mathcal{M}$, batch size $b$, inner-loop learning rate $\alpha$, test set $T$. }
	\KwOut{Predictions on the test set}
	$S \leftarrow \mathrm{sample} (\mathcal{M}, m \cdot b)$ \\
	$Q \leftarrow T$ \\
	$\boldsymbol{w}' = \mathrm{SGD}( \mathcal{L},  \boldsymbol{\phi}_e, \boldsymbol{\phi}_{proto}, \boldsymbol{w}, S, \alpha)$  \\
	$\rm{predict} (Q, \boldsymbol{\phi}_e, \boldsymbol{\phi}_{proto}, \boldsymbol{w})$ 
\end{algorithm}

\subsection{Prototype-guided Memory Replay Module}\label{subsec: proto-memory}

We aim to leverage a small amount of previously seen examples to retain the knowledge learned from the preceding tasks. A prototypical network provides a solution to characterize feature representations by prototypes. Thereby, we propose a replay memory module to select and revisit worthwhile samples by means of a prototypical network.

\subsubsection{Prototypical Network} We devise a prototype-guided memory replay module to selectively write examples from query set into memory $\mathcal{M}$, by using prototypical knowledge. We integrate a prototypical network \cite{Snell2017PrototypicalNF} into our model and establish a few-shot learning framework inside prototypical learning process. It is formulated as follows, given a small support set of $M$ labeled samples $S = {(x_1, y_1),...,(x_M, y_M)}$ where each $x_i$ is a $D$-dimensional feature vector of an example as $x_i \in \mathbb{R}^D$ with the corresponding label $y_i \in {1,...,L}$. $S_l$ is the support set of examples of class $l$. $Q_l$ is the query set of examples of class $l$. In prototypical network, $S_{L}$ denotes the overall support set for all classes $S_{L}$ and $Q_{L}$ denotes the overall query set.

In particular, we use a prototypical network to learn an $M$-dimensional representation $\boldsymbol{c}_l \in \mathbb{R}^{M}$, a prototype for class $l$. The prototype of each class is derived by the embedding function $h_{\boldsymbol{\phi}}: \mathbb{R}^{D} \rightarrow \mathbb{R}^{M}$ in representation learning network. In class $l$, the prototype is computed as the mean vector of the embedded support samples, that is:
\begin{equation}
    \mathbf{c}_l \leftarrow \frac{1}{|S_l|} \sum \limits_{\substack{ (x_i, y_i) \in S_l}} h_{\boldsymbol{\phi}_{proto}}(x_i)
\end{equation}

The learning process involves minimising a negative log-probability of the real class $l$, that is $J(\boldsymbol{\phi}_{proto}) = -\log p_{\boldsymbol{\phi}_{proto}} (y = l | x)$. For a training episode, we randomly obtain a subset of classes from the given training samples. In this case, the meta learning support set $S_i$ serves as the training set for learning a proper prototypical function. Based on the selected subset of classes, we randomly samples a subset of data within each class and forms a support set. $S_l$ denotes the support set for class $l$. The query set is the subset of the reminder and $Q_l$ denotes the query set for class $l$. The parameters $\boldsymbol{\phi}_{proto}$ is finetuned using query set $Q_{L}$. The learning process perform SGD with the learning rate $\alpha$. 

Each meta-learned prototypical vector in $\{\boldsymbol{c}_1,...,\boldsymbol{c}_L \}$ serve as a reference to acquire samples that are representative for the correspond class. Subsequently, we employ the $K$-Nearest Neighbours approach over the meta learning query set to acquire the closest data points to the prototypical vectors and save into memory $\mathcal{M}$ for episodic replay.

\begin{algorithm}[t]
	\caption{writeSamples$(\mathcal{M}, Q_{i},\boldsymbol{c}_l, n)$}
	\label{algo: write_func}
	\KwIn{Memory $\mathcal{M}$, query set $Q_{i}$, prototype $\boldsymbol{c}_l$, number of selected samples per class $n$.}
	\KwOut{Updated Memory $\mathcal{M}$}
	\eIf{\rm{samples} of \rm{class} $k$ \rm{not in} $\mathcal{M}$} {
		$X_l$ $\leftarrow$ \rm{KNearestNeighbour} ($Q_i, \boldsymbol{c}_l, K = n$)  }{ 
		$X_l$ $\leftarrow$ \rm{KNearestNeighbour} ($X_l \cup Q_i, \boldsymbol{c}_l, K = n$)
		}
	write $X_l$ into $\mathcal{M}$  \tcp*{ Sample set $X_l$ contains $n$ selected data points for class $l$}

\end{algorithm}

\subsubsection{Dynamic Memory}
Our devised memory is dynamically updated, considering the renewal of the prototype for each class in each training epoch. Given the updated prototypes, our memory module leverages $K$-Nearest Neighbour. It chooses $K$ samples that closest to the prototype for each class from both query set $Q_i$ and the previously stored samples. Hereby, the memory size growing linearly with the number of trained classes. In this paper, we opt for 5 nearest neighbours. Particularly, we assign $n$ as the number of selected samples per class to 5 in Algorithm \ref{algo: write_func}.  Replay sample selection merely leverage the up-to-date prototypes. The write mechanism are displayed in Algorithm \ref{algo: write_func}.

\subsubsection{Replay Frequency \& Sample Replay Scheme} In most replay-based approaches, it is sufficient to set the replay frequency as 1\% to adapt to new tasks. Under such a setup, some works  \cite{wang-etal-2020-efficient, Huang2021ContinualLF} show that a random approach to retrieve previous training samples is the most efficient strategy. Without consuming much computational power, the model performance of using random samples selection is quite close to the best. Our data selection scheme is prone to acquiring representative examples. While, revisiting representative samples from old tasks too frequently can cause a problem while learning a new task. It disrupts the learning of underlying novel data distribution. Thereby, we adjust the episodic replay frequency to better fit an efficient learning process. In this paper, we activate reading from memory $\mathcal{M}$ every 50 epochs in each learning task. Replay frequency is computed according to different input sequences, shown in Table ~\ref{tab:replay-rate}. In our model, the frequency changes along with the number of the seen classes. The replay rate for samples per class is approximately 0.0667 \%.

\section{Experiments} \label{sec:experiments}
\subsection{Datasets}
For performance evaluation, we conduct experiments on text classification in various training set orders. Following prior work on class-incremental learning, we use the benchmark datasets initially introduced by \cite{dAutume2019EpisodicMI}. We conduct the experiments on three datasets, i.e., AGNews (news classification; 4 classes), Yelp (sentiment analysis; 5 classes) and Amazon (sentiment analysis; 5 classes). Each dataset contains 115,000 training samples and 7,600 test samples. Among them, Yelp and Amazon datasets have similar semantics (i.e., product rating), which share the same label spaces. In the experiments, we train our model and baselines on these datasets sequentially where each dataset is considered as a separated learning task. There are a total of 6 different ways to train the three datasets sequentially, shown in Table \ref{tab:sequence orders}. Hereby, we evaluate our model performance on these datasets in 6 various orders.

\begin{table}[!t]
\renewcommand{\arraystretch}{1.3}
\caption{\label{tab:sequence orders} The Input Datasets Orders}
\centering
\begin{tabular}{|c||c|}
\hline
\textbf{No.} & \textbf{Dataset Orders} \\
\hline
1 & Yelp $\rightarrow$ AGNews $\rightarrow$ Amazon  \\
2 & Yelp $\rightarrow$  Amazon $\rightarrow$ AGNews\\
3 &  Amazon $\rightarrow$ Yelp $\rightarrow$ AGNews \\
4 &  Amazon $\rightarrow$ AGNews $\rightarrow$ Yelp \\
5 & AGNews $\rightarrow$ Yelp $\rightarrow$ Amazon \\
6 & AGNews $\rightarrow$ Amazon $\rightarrow$ Yelp \\ 
\hline
\end{tabular}
\end{table}

\begin{table*}[!t]
\renewcommand{\arraystretch}{1.3}
\caption{\label{tab:replay-rate} The Replay Rates of CL Methods in Six Dataset Orders}
\centering
\begin{tabular}{|l||ccc|}
\hline
\multirow{2}*{\textbf{Method}}  & \multicolumn{3}{c|}{\textbf{Order 1 \& 4}} \\
& Yelp (or Amazon)  & AGNews & Amazon (or Yelp) \\
\hline 
PMR$_{argmin}$  &  \textbf{0.3}\% & \textbf{0.7}\% & \textbf{0.6}\% \\
PMR$_{argmin,1\%}$ &  1.00 \% & 1.00\%  & 1.00\% \\
\hline
Other Baselines  &  1.00 \% & 1.00\%  & 1.00\% \\
\hline
\hline
\multirow{2}*{\textbf{Method}} & \multicolumn{3}{c|}{\textbf{Order 2 \&  3}} \\
& Yelp (or Amazon) & Amazon (or Yelp)  & AGNews \\
\hline 
PMR$_{argmin}$  & \textbf{0.3}\% & \textbf{0.3}\% & \textbf{0.7}\% \\
PMR$_{argmin,1\%}$ &  1.00 \% & 1.00\%  & 1.00\% \\
\hline
Other Baselines  &  1.00 \% & 1.00\%  & 1.00\% \\
\hline
\hline
\multirow{2}*{\textbf{Method}} & \multicolumn{3}{c|}{\textbf{Order 5 \& 6}} \\
 & AGNews & Yelp (or Amazon) & Amazon (or Yelp)  \\
\hline 
PMR$_{argmin}$  & \textbf{0.3}\% & \textbf{0.6}\%   &  \textbf{0.6}\%  \\
PMR$_{argmin,1\%}$ &  1.00 \% & 1.00\%  & 1.00\% \\
\hline
Other Baselines  &  1.00 \% & 1.00\%  & 1.00\% \\
\hline
\end{tabular}
\end{table*}

\subsection{Setup}
Unlike the experimental setup of prior works, we consider a low resource setup, in which we constrain the memory size to 45 samples (i.e., 5 samples per classes in our model). We prune the input sequence length to 200, and use a pretrained ALBERT-Base-v2 from Hugging Face Transformers \cite{DBLP:journals/corr/abs-1909-11942} as the encoder for our model and all baselines. The write rate for all baselines is set to 1. OML-ER \cite{DBLP:journals/corr/abs-2009-04891}, ANML-ER \cite{DBLP:journals/corr/abs-2009-04891} and PMR employ meta learning in the training process. For their meta-learning setup, we employ SGD in the inner loop optimization with learning rate, $\alpha = 3e^{-3}$ and Adam in the outer loop optimization with learning rate, $\beta = 3e^{-5}$. A-GEM and Replay model use Adam as their optimizer with learning rate $3e^{-5}$. A-GEM \cite{AGEM} , Replay \cite{DBLP:journals/corr/abs-2009-04891}, OML-ER and ANML-ER utilise the random sampler, which samples data in a stochastic manner with batch size, $b$ = 25. Our example sampler randomly selects five training samples from different classes for each epoch without replacement. The batch size depends on the number of classes from the training dataset. In the experiments, there are two batch sizes for each task, i.e., $b$ = 20 for AGNews and $b$ = 25 for Yelp and Amazon. The number of mini-batches in each epoch, $m$ = 5.  As for replay frequency, all baselines follow sparse experience replay rate proposed by \cite{dAutume2019EpisodicMI}. That is retrieving 1\% samples from witnessed samples for learning. Specifically, 150 samples are revisited for every 15,000 samples. In PMR, we assign the frequency to every 50 epochs, resulting in different replay rates, shown in Table ~\ref{tab:replay-rate}.

For evaluation, test sets order is the same as training sets order. In meta testing shown in Algorithm \ref{algo:testing}, each test set is established as one episode for each of the learned task. Specifically, the query set is constructed by the test set. Hereby, we evaluate model performance on the query set. We record the average result of 3 independent runs. All models are executed on a Linux platform with 8 Nvidia Tesla A100 GPU and 40 GB of RAM. All experiments are performed using PyTorch \cite{NEURIPS2019_9015}.

\subsection{Baselines}
In our experiments, we utilize the following CL baselines for comparisons.

\begin{itemize}
    
    \item \textbf{A-GEM} \cite{AGEM} is a commonly-used CL baseline. It imposes one gradient constraint to restrict current task gradient projection regions. A-GEM randomly acquires examples from a preceding task data buffer to outline the direction of optimization constraints. A-GEM has the same experience replay setting as OML-ER.

    \item \textbf{Replay} model simply incorporates sparse experience replay into a sequential learning model. Replay model utilises the same experience replay strategy as OML-ER. It performs one gradient update on randomly drawn replay samples.

    \item \textbf{OML-ER} \cite{DBLP:journals/corr/abs-2009-04891} is the latest state-of-the-art CL model in class-incremental learning that applying online meta learning (OML) \cite{Javed2019MetaLearningRF} model with episodic experience replay. In experiments, we limit the memory budgets to 45 samples. OML-ER utilises random selection scheme in both read and write mechanisms.
    
    \item \textbf{ANML-ER} \cite{DBLP:journals/corr/abs-2009-04891} is also a strong baseline, that outperforms OML-ER in some cases. It combines neuromodulated meta learning (ANML) \cite{DBLP:conf/ecai/BeaulieuFMLSCC20} model with episodic experience replay. In experiments, ANML-ER performs incompetently while memory size reducing to 45. Accordingly, we only limit the size of replay samples from the support set to 45. That is 57,070 stored replayed samples in total, which downsizing by around 83.25 \% for data storage. Its replay frequency and replay sample size are the same as all other baselines.
    
    \item \textbf{PMR$_{argmin,1\%}$} is PMR with 1\% experience replay rate. For PMR, the number of revisited samples in each learning task is decided by the number of learned classes $|\mathcal{M}|$ and batch size $b$ in the experiments. Thereby, we adjust the replay frequency to harness the replay rate. Concretely, the proposed model, PMR$_{argmin}$ revisits $|\mathcal{M}|$ samples every $b\cdot (m+1) \cdot R_F + b\cdot m$ samples. Whereas, PMR$_{argmin,1\%}$ sustains a 1\% replay rate.
\end{itemize}

\begin{table*}[t!]
\renewcommand{\arraystretch}{1.3}
\caption{\label{tab:accuracy} Evaluation on the Performance in Different Training Set Orders in terms of Accuracy}
\centering
\begin{tabular}{|l||cccccc||l|}
\hline
\textbf{Method} & Order 1 & Order 2 & Order 3 & Order 4 & Order 5 & Order 6 & Average \\
\hline 
\hline
AGEM & 37.62 & 30.20 & 30.55 & 41.94 &  39.59 & 39.75 & $36.61 \pm 5.02 $\\
Replay & 43.76  & 30.07 &  30.45  & 41.91 & 42.12 & 44.42 & $38.79 \pm 6.68$\\
OML-ER &  45.73 &  \textbf{46.44} & 41.68 & 47.49 & \textbf{60.83} & 62.11 & $50.71 \pm 8.57$ \\
ANML-ER & 30.83  & 23.21 &  20.22 & 40.39 & 29.82 & 33.66 & $29.69 \pm 7.25$ \\
\hline
PMR$_{argmin}$  & 59.46 &  38.06 & 39.70 & 56.88 & 59.31 &\textbf{62.19} & $ 52.60 \pm 10.77 $ \\
PMR$_{argmin,1\%}$   &  \textbf{60.23}  & 39.81 & \textbf{41.12} &  \textbf{58.24} & 59.29 & 58.02 & $ \mathbf{52.79 \pm 9.59} $ \\

\hline
\end{tabular}
\end{table*}

\begin{figure*}[!t]
\centering
\subfloat[Order 1]{\includegraphics[width=2.3in]{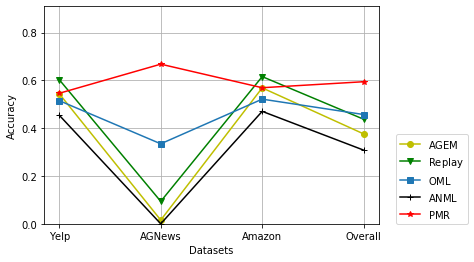}
\label{fig:201}}
\hfil
\subfloat[Order 2]{\includegraphics[width=2.3in]{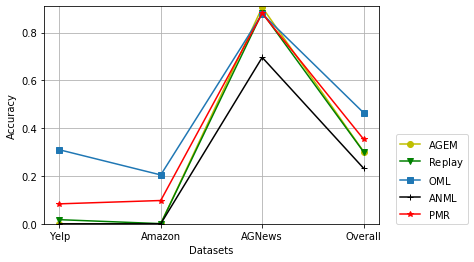}
\label{fig:210}}
\hfil
\subfloat[Order 3]{\includegraphics[width=2.3in]{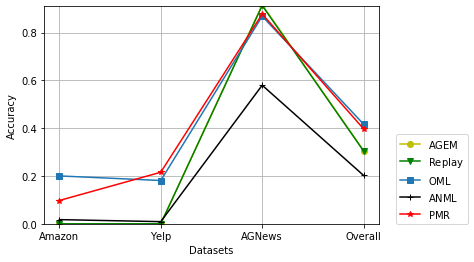}
\label{fig:120}}
\hfil
\subfloat[Order 4]{\includegraphics[width=2.3in]{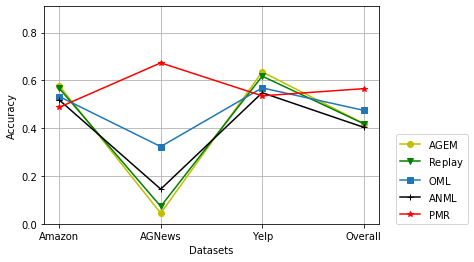}
\label{fig:102}}
\hfil
\subfloat[Order 5]{\includegraphics[width=2.3in]{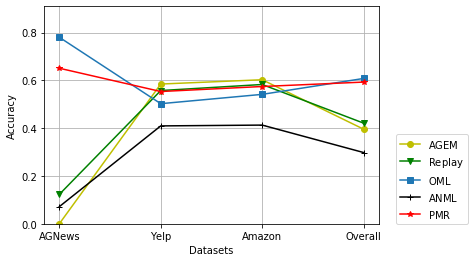}
\label{fig:021}}
\hfil
\subfloat[Order 6]{\includegraphics[width=2.3in]{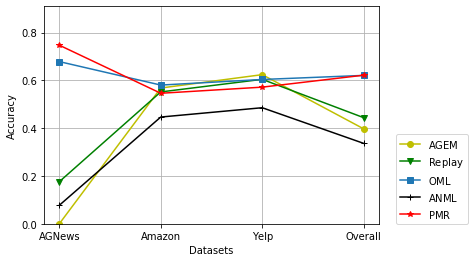}
\label{fig:012}}
\caption{\label{fig:order}Performance evaluations on test examples from each learned dataset. Note that X-axis is corresponding to the dataset orders in evaluation, as in training. PMR in all subfigures denotes the proposed model, PMR$_{argmin}$. OML and ANML denotes OML-ER and ANML-ER, respectively.}
\end{figure*}

\subsection{Results}
We examine model performance on text classification in terms of test set accuracy in 6 different input sequences. For evaluation, we consider model performance on a test set for each of the tasks $\mathcal{T}$ after terminating the learning of the last task $\mathcal{T}^{(K)}$. Concretely, letting $A_{K,k}$ be the test classification accuracy of the model on task $\mathcal{T}^{(k)}$ after completing the learning of the last task $\mathcal{T}^{(K)}$, we define the evaluation metric, average accuracy as:

\begin{equation}
    \mathrm{ACC} = \frac{1}{K} \sum^{K}_{k=1} A_{K,k}
\end{equation} In Table ~\ref{tab:accuracy}, we present the average accuracy across baselines and our model with standard deviations across all 6 sequences in the last column.

For various orders, both of the PMR models have an outstanding performance. PMR$_{argmin,1\%}$ achieves the highest accuracy in order 1, order 3 and order 4. While PMR$_{argmin}$ achieves the highest accuracy order 6. PMR$_{argmin,1\%}$ with a higher replay rate leads PMR$_{argmin}$ in overall performance only by 0.19\%.  Whereas, PMR$_{argmin}$ is the second best model, and outperforms the latest SOTA, OML-ER, by approximately 1.89\% higher. We argue that PMR$_{argmin}$, is the most efficient approach compared to other baselines. OML-ER shows its competitiveness and obtain the highest accuracy in order 2 and order 5. In this experiment, only PMRs and OML-ER achieve the averaged accuracy higher than 50\%. The performance of Replay model is better than A-GEM and ANML-ER. A-GEM is inferior to Replay method, where its overall accuracy is lower by around 2.18\%. Similar results from other literature \cite{DBLP:journals/corr/abs-2009-04891, wang-etal-2020-efficient} suggest constraint in the gradient space for knowledge consolidation may not be as good as simply revisiting samples. The evaluation outcome of ANML-ER is surprisingly bad. The reason might be its inflexibility towards our constraints on memory budgets. 

It is evident that the performance of all baselines and our model are sensitive to training set orders. Concretely, all examined models perform poorly in order 2 and order 3. A-GEM presents the most stable performance across all sequences, with a standard deviation of 5.02\%. Replay, OML-ER and ANML-ER have a standard deviation below 10\%. However, the top 2 well-performed models, PMRs, produce a standard deviation larger than 10\%. These outcomes indicate that a well-performed CL model is likely to cause a large performance variation in various training set orders. More details are elaborated in Sec.~\ref{sub:analysis}.

\subsection{Analysis} \label{sub:analysis}
\subsubsection{Effect of Training Sets Orders} We further study the effect of training sets orders in terms of accuracy on all tasks. We illustrate the test performance of our model on the test sets of Yelp, AGNews and Amazon respectively in Fig.~\ref{fig:order}. The overall accuracy is the average result obtained from three learned tasks. 

In Fig.~\ref{fig:201} and Fig.~\ref{fig:102}, all baselines exhibit a sudden accuracy drop on AGNews, while their performances on Yelp and Amazon are relatively good. To be specific, all baselines are able to learn the distribution from Yelp and Amazon well. This is because that these datasets are the latest learned task or share the same domain with the latest learned task. Whereas, all baselines experience a severe forgetting on the second learned task, AGNews, in which their accuracy is lower than 20 \%. OML-ER attains higher accuracy on AGNews compared to other baselines. But the accuracy drops at least 10 \%, compared to learning Yelp and Amazon. As for PMR, not only it has a competitive performance on Yelp and Amazon, also PMR has a surprisingly good result on AGNews, in which its accuracy is around 33 \% higher than the second best model OML-ER. The result demonstrates our model has an outstanding sequential learning ability in Order 1 and Order 4, which implies its good tolerance of catastrophic forgetting.

In Fig.~\ref{fig:210} and Fig.~\ref{fig:120}, the overall experimental results of all methods are not good. Even though the first and second learned tasks share the same domain and hereby have 2 times larger training examples, the ultimately learned distribution is heavily skewed towards the last learned task. The accuracy of A-GEM, Replay and ANML-ER on Yelp and Amazon declines to almost 0. While, OML-ER and PMR have better performances, seeing that their accuracy on the first two preceding tasks is not dropped to 0. In meta-testing, both OML-ER and PMR are fine-tuned on test samples in the inner-loop optimization. We argue that task-specific fine-tuning in the inference stage prevents severe catastrophic forgetting. However, PMR and OML-ER are incompetent in text classification on Yelp and Amazon.

\begin{table*}[t!]
\renewcommand{\arraystretch}{1.3}
\caption{\label{tab:ablation} Ablation Study on Replay Sample Selection Strategies}
\centering
\begin{tabular}{|l||ccc||l|}
\hline
\textbf{Method} & Yelp & AGNews & Amazon & Average \\
\hline 
\hline
OML-ER  & 51.53  & 33.48 & 52.17 & $45.73 \pm 10.61 $ \\
\hline

PMR$_{augment}$  & 53.69  & 48.02 & 54.97 & $52.23 \pm 3.70$ \\

PMR$_{argmax}$  & 49.73  & 37.50 & 51.72  & $46.32 \pm 7.70$\\

PMR$_{mix}$  &  \textbf{56.60} & 58.14 & \textbf{57.26} & $57.33 \pm 0.77$ \\
\hline
\textbf{PMR$_{argmin}$}  & 54.63 & \textbf{66.81} & 56.95 & $\mathbf{59.46 \pm 6.47}$ \\
\hline
\end{tabular}
\end{table*}

\begin{table*}[t!]
\renewcommand{\arraystretch}{1.3}
\caption{\label{tab:forget}Ablation Study on Forgetting.}
 \centering
\begin{tabular}{|l||ccc|}
\hline
\multirow{2}*{\textbf{Method}}  & \multicolumn{3}{c|}{\textbf{Yelp}} \\
& Single Task Learning & Sequential Learning & Accuracy Drop \\
\hline 
\hline
OML-ER &  58.40 & 51.53 &  6.87 \\
\hline
PMR$_{augment}$  & 58.19  & 53.69 & 4.50 \\

PMR$_{argmax}$  &  53.74 &  49.73  & 4.01 \\

PMR$_{mix}$  & 57.05  &  \textbf{56.60} & \textbf{0.45} \\

\textbf{PMR$_{argmin}$}  &  \textbf{58.68} & 54.63 & 4.05 \\
\hline
\multirow{2}*{\textbf{Method}} & \multicolumn{3}{c|}{\textbf{AGNews}} \\
& Single Task Learning & Sequential Learning & Accuracy Drop \\
\hline 
\hline
OML-ER  &  86.82 & 33.48  & 53.34 \\
\hline
PMR$_{augment}$  &  89.76 & 48.02 &  41.74 \\

PMR$_{argmax}$  & \textbf{88.89}  & 37.50 & 51.39 \\

PMR$_{mix}$  & 88.68  & 58.14  & 30.54 \\

\textbf{PMR$_{argmin}$}  & 87.43  & \textbf{66.81} & \textbf{20.62} \\
\hline
\multirow{2}*{\textbf{Method}} & \multicolumn{3}{c|}{\textbf{Amazon}} \\
& Single Task Learning & Sequential Learning & Accuracy Drop \\
\hline 
\hline
OML-ER  & 44.94 & 52.17  & -7.23 \\
\hline
PMR$_{augment}$  &  \textbf{56.63} & 54.97 & 1.66 \\

PMR$_{argmax}$  &  49.97 &  51.72 &  -1.75 \\

PMR$_{mix}$  & 53.08  & \textbf{57.26} & -4.18  \\

\textbf{PMR$_{argmin}$}   & 49.01  & 56.95  & \textbf{-7.94}  \\
\hline
\end{tabular}
\end{table*}

In Fig.~\ref{fig:021} and Fig.~\ref{fig:012}, A-GEM, Replay and ANML-ER demonstrate similar learning behaviors in Fig.~\ref{fig:201} to Fig.~\ref{fig:120}. These models perform well only in the domain of the last learned task. In these models, catastrophic forgetting always occurs on the preceding learned tasks. Both OML-ER and PMR show a better performance on AGNews rather than Yelp and Amazon. The outcome of PMR is foreseeable by its performance shown in Fig.~\ref{fig:201} and Fig.~\ref{fig:102}. However, it is unexpected to see how well OML-ER behaves on AGNews in Order 5 and Order 6. Test set orders may be the potential reason, since OML-ER is fine-tuned on AGNews first at inference stage.

From Fig.~\ref{fig:201} to Fig.~\ref{fig:012},  we spot a phenomenon regarding PMR's learning behaviour. In sequential learning Yelp and Amazon, no matter the orders of these two tasks,  PMR always returns a better accuracy on the later learned task than the preceding one. Therefore, we argue that PMR exhibits learning performance improvement in a similar task from the same domain, compared to other baselines.

\begin{figure*}[!t]
\centering
\subfloat[Unigram Distribution of Saved Samples in Episode 1]{\includegraphics[width=7.2in]{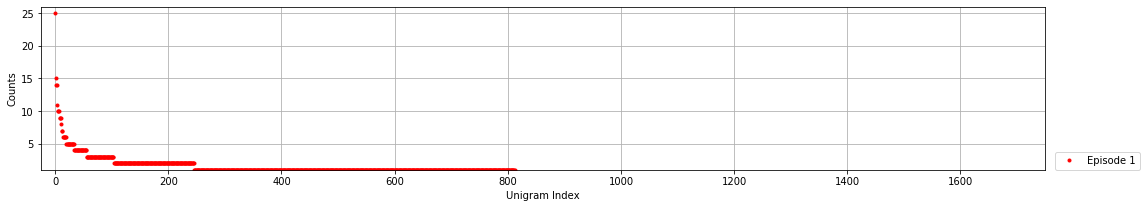}
\label{fig:1}}
\hfil
\subfloat[Unigram Distribution of Saved Samples in Episode 408]{\includegraphics[width=7.2in]{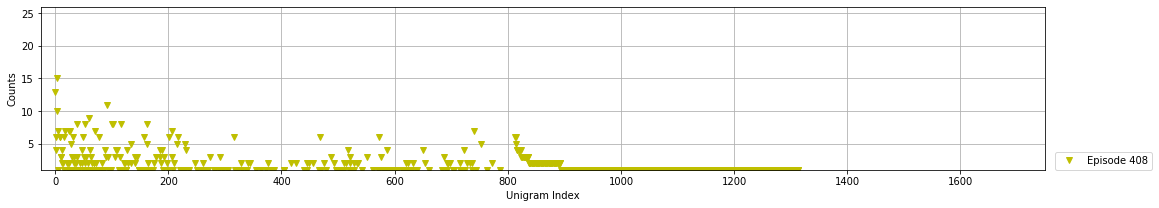}
\label{fig:408}}
\hfil
\subfloat[Unigram Distribution of Saved Samples in Episode 765]{\includegraphics[width=7.2in]{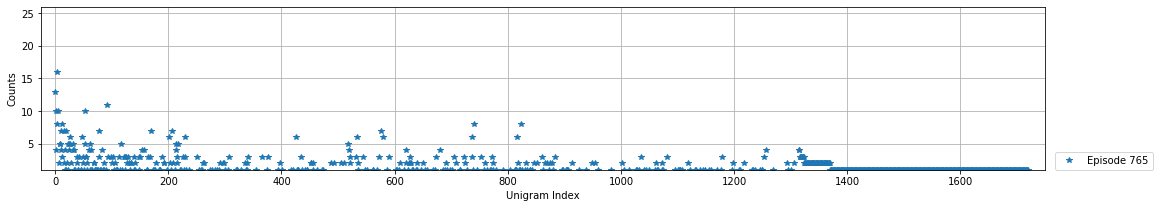}
\label{fig:765}}
\hfil
\subfloat[Comparison of Unigram Distributions in Episode 1, Episode 408 and Episode 765]{\includegraphics[width=7.2in]{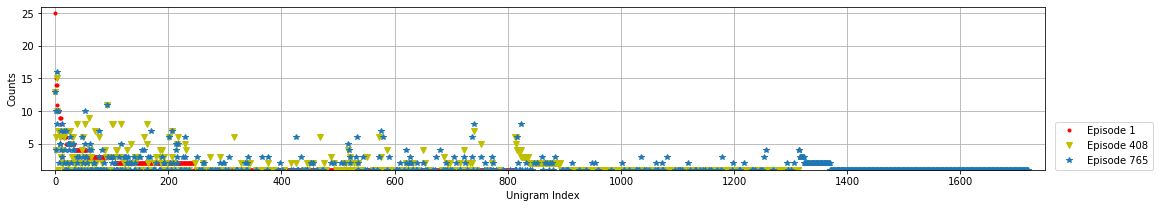}
\label{fig:all}}
\caption{\label{fig:insight} Visualization of Unigram Distribution Shift in Memory. Note that y-axis is in the range $\boldsymbol{[1,26)}$. The data points on the x-axis indicate the count of the corresponding unigram is 1.}
\end{figure*}

\subsubsection{Effect of Replay Sample Selection} We perform an ablation study on different replay sample selection schemes. In the proposed model, each prototype represents a learned class and steers samples selection for memory replay. Whereas, the strategies of choosing data point via prototypes can be different. The following 4 strategies are examined.

\begin{itemize} 
    \item \textbf{PMR$_{\boldsymbol{argmin}}$} leverages the proposed replay sample selection approach. It finds 5 representative samples from the \textit{query set} for each class in each training epoch. Particularly, it chooses five nearest examples to each prototype by euclidean distance.

    \item \textbf{PMR$_{\boldsymbol{augment}}$} is the extension of PMR$_{argmin}$. The sample selection set in each training epoch includes not only the \textit{query set}, but also the \textit{support set}. Thus, it picks 5 representative samples from a five-times larger samples selection set compared to PMR$_{argmin}$.

    \item \textbf{PMR$_{\boldsymbol{argmax}}$} leverages 5 outliers of each class for experience replay, by finding the samples with maximum euclidean distances to the prototype.

    \item \textbf{PMR$_{\boldsymbol{mix}}$} combines the sample selection approaches of PMR$_{argmin}$ and PMR$_{argmax}$. In particular, it replays 5 representative samples and 5 outliers from each class. The outliers are only applied in the current learned tasks. It will be abandoned once the training of the current task ends.
\end{itemize}

We evaluate these PMR models along with the strong baseline, OML-ER on the datasets in Order 1.

The experimental results are shown in Table \ref{tab:ablation}. The last column is the average accuracy with standard deviations across all three learned tasks. It is predictable that PMR$_{argmax}$ shows the worst performance on all three learned task and the average accuracy. Outliers are not valuable retrieve information for learning. PMR$_{augment}$ is the second worst model, but better than the baseline OML-ER. Replaying the most representative samples loses generalization to some extent, resulting in performance degradation. PMR$_{mix}$ attains the highest accuracy on Yelp and Amazon. The impact of representative samples from the current task are neutralized by its outliers, given more attention to samples from preceding tasks in each review of learned knowledge. Though PMR$_{mix}$ performs better on Yelp and Amazon, it only improves the performance on Amazon by less than 1 \%. While, PMR$_{argmin}$ shows a more than 2\% accuracy increase on Amazon. Additionally, it obtains the highest accuracy on AGNews and achieves the highest overall accuracy. PMR$_{argmin}$ finds the representative samples only from query set, resulting in a better performance than PMR$_{augment}$. PMR$_{augment}$ chooses samples from both the support set and the query set, indicating the selected samples are more task specific compared to PMR$_{argmin}$. It suggests that revisiting relatively generalized samples are better than retrieving more task-specific samples. Furthermore, except PMR$_{argmax}$, all PMR models outperform than OML-ER. It can be seen that sample selection matters. Various sample selection strategies cause different model performance. We find out the good selection strategies, that use prototypical information as the reference. Arguably, the proposed prototype-guided approach is better approach to sample selection for experience replay, compared to commonly-used random selection schemes.

\subsubsection{Forgetting Evaluation}To further investigate the effect of various replay sample selection strategies,we conduct an ablation study with regard to forgetting.

We evaluate model performance in a single task learning manner and a sequential multi-task learning manner (i.e., the CL manner). In this experiment, the evaluation metric of forgetting is accuracy drop between single-task learning and sequential learning. A smaller value of accuracy drop indicates less forgetting on a learned task. As shown in Table \ref{tab:forget}, for the first learned task Yelp, PMR$_{mix}$ shows the least forgetting of the task. PMR$_{argmin}$ also demonstrates its good ability of knowledge consolidation. Its accuracy drop in the second learned task is 20.62 \%, which is lower than the second-best model by almost 10 \%. Whereas, OML-ER exhibits the worst memorization of the first and the second learned task, which causes a degrading of its performance by 6.87\% and 53.34 \% respectively. Seeing that all models sequentially learned Yelp before Amazon, a forward knowledge transfer is presented in OML-ER, PMR$_{argmax}$, PMR$_{mix}$ and PMR$_{argmin}$. Instead of accuracy drop, the accuracy of these models on Amazon is improved by 1.94 \% to 7.94 \%. Among them, PMR$_{argmin}$ achieves the largest improvement while OML-ER is the second-best in learning ability improvement. It can be seen that PMR$_{argmin}$ enables efficient knowledge transfer between tasks from the same domain. Also, this experiment once again verified the effect of PMR$_{argmin}$ on catastrophic forgetting.

\subsubsection{Memory Insight}
To study whether the synthetic prototypes are constantly updated to realize sample efficiency in the selection, we visualize unigram distribution shift in memory in three different learning phases, i.e., Episode 1, Episode 408 and Episode 765. The experiment is conducted on Yelp to analyze the performance of selection strategy in PMR$_{argmin}$. We fix the memory size to 25 examples and examine the sample efficiency by the unigram distribution of these saved examples after the completion of writing in memory. The experimental results are shown in Fig. \ref{fig:insight}. The y-axis presents the counts of each unigram. While, for simplicity, the x-axis presents the unigram index instead of the unigram. The evaluation is performed in three different learning phases and presented in Fig. \ref{fig:1}, Fig.\ref{fig:408} and Fig.\ref{fig:765} respectively. The variety of unigrams in all three learning phases is 1,720 in total. For comparison, Fig. \ref{fig:all} demonstrates the change of unigram distribution in different learning episodes.  

Fig. \ref{fig:1} shows the first learning phase, which writes 25 samples in memory selected by the initially learned prototypes. It is obvious that the unigrams from the saved samples in Episode 1 distribute merely on the first halves of the x-axis. The number of unigrams with only 1 appearance is 567. The distribution is heavily skewed towards certain values, where an extreme value 25 appears in the unigram indexed as 0. Arguably, the selected samples fail to contain the diverse feature representations. PMR$_{argmin}$ performs 15 experience replays in the learning process to enhance learned knowledge. Therefore, we examine the stored examples for the 8th experience replay in Episode 408 and the last experience replay in Episode 765. In Fig. \ref{fig:408}, the unigram distribution indicates the saved samples in Episode 408 cover a wider range of unigrams compared to that in Episode 1. The selected samples not only contain the newly seen unigrams, but also some of the unigrams shown in Episode 1. It exhibits the improvement in sample selection, considering that PMR$_{argmin}$ continually adjusts its selection strategy via up-to-date prototypes. Fig. \ref{fig:765} visualizes the unigram distribution of the samples for the last experience replay in Episode 765. Compared to the previous two experimental results, the samples in Episode 765 hold the unigrams in more broad distribution. It implies a good diversity of information given by the saved samples, even though the amount of the samples is limited. Similarly, it preserves some of the previously attained unigrams and introduces newly seen unigrams. Some of the old unigrams are emphasised while some are neglected. It demonstrates the adjustment of the chosen feature representations in the learning process. Consequently, we testify that the prototypical-guided selection strategy enables samples efficiency, thereby reducing memory budget.

\section{Conclusion} \label{sec:conclusion}

To summarise, we utilise a prototypical network in CL for knowledge consolidation, hereby alleviating catastrophic forgetting. Particularly, the prototypes constantly learn from the seen training samples to guide the selection of a limited number of samples for experience replay. A dynamically updated memory module is devised with low resource consumption. By strictly limiting the memory size and replay rate, we lessen the memory budget by achieving sample efficiency. 

The proposed model shows its significant performance in avoiding catastrophic forgetting and enabling efficient knowledge transfer in the experiments. It exhibits the feasibility and validity of using the prototypical network in CL approaches. With a less than 1\% replay rate, PMR$_{argmin}$ still outperforms CL baselines in terms of accuracy. It demonstrates its efficiency in memory resource allocation. Whereas, we show that training set orders can influence CL model performance. Therefore, we plan to further improve our model's stability on different training sets orders. Additionally, we intend to optimize resource allocation in PMR$_{argmin}$ to realize low resource consumption. Furthermore, existing CL models mainly operates in a centralized setting. CL as an efficient knowledge transfers approach leverages a relatively small amount of training samples, which can be seen as a promising solution to address learning in a decentralized setting. A future research direction can be extending the proposed model into a decentralized research scenario.

\ifCLASSOPTIONcaptionsoff
  \newpage
\fi



\bibliographystyle{IEEEtran}
\bibliography{arxiv}
%



%

\end{document}